\documentclass[final]{cvpr}

\usepackage{times}
\usepackage{epsfig}
\usepackage{graphicx}
\usepackage{amsmath}
\usepackage{amssymb}

\usepackage{multirow}
\usepackage{booktabs}

\usepackage[pagebackref=true,breaklinks=true,colorlinks,bookmarks=false]{hyperref}

\def\cvprPaperID{2244} 
\def\confYear{CVPR 2021}

\begin{document}

\title{Learning to Generate Content-Aware Dynamic Detectors}

\author{Junyi Feng$^{1,2}$, Jiashen Hua$^{2}$, 
Baisheng Lai$^{2}$, Jianqiang Huang$^{2}$,
Xi Li$^{1}$, Xian-sheng Hua$^{2}$\\
$ ^1$Zhejiang University, $ ^2$Damo Academy, Alibaba Group\\
{\tt\small \{fengjunyi, xilizju\}@zju.edu.cn,}\\
{\tt\small \{jiashen.hjs, baisheng.lbs, jianqiang.hj, xiansheng.hxs\}@alibaba-inc.com}
}

\maketitle

\begin{abstract}
  Model efficiency is crucial for object detection. Most previous works rely on either hand-crafted design or auto-search methods to obtain a static architecture, regardless of the difference of inputs. In this paper, we introduce a new perspective of designing efficient detectors, which is automatically generating sample-adaptive model architecture on the fly. The proposed method is named content-aware dynamic detectors (CADDet). It first applies a multi-scale densely connected network with dynamic routing as the supernet. Furthermore, we introduce a course-to-fine strategy tailored for object detection to guide the learning of dynamic routing, which contains two metrics: 1) dynamic global budget constraint assigns data-dependent expected budgets for individual samples; 2) local path similarity regularization aims to generate more diverse routing paths. With these, our method achieves higher computational efficiency while maintaining good performance. To the best of our knowledge, our CADDet is the first work to introduce dynamic routing mechanism in object detection. Experiments on MS COCO dataset demonstrate that CADDet achieves 1.8 higher mAP with 10\% fewer FLOPs compared with vanilla routing strategy. Compared with the models based upon similar building blocks, CADDet achieves a 42\% FLOPs reduction with a competitive mAP.
\end{abstract}
\section{Introduction}
\label{intro}

Object detection, one of the core tasks in computer vision, aims at localizing and classifying objects in an image.
In the past few years, various object detection models~\cite{fast-rcnn, faster-rcnn, fpn, cascade-rcnn, yolo, yolo9000, ssd, retinanet, centernet, fcos}
based on deep convolutional neural networks have been proposed.
A general object detector consists of three modules, including a backbone, a neck, and a detection head. 
To achieve higher performance, all the modules should be designed delicately.
Recently, Neural Architecture Search~(NAS) strategies have demonstrated promising results on the object detection 
task~\cite{nas-fpn, mnas-fpn, nas-fcos, efficientdet, spinenet, sp-nas, detnas, hit-det}.
Most of these works~\cite{nas-fpn, mnas-fpn, nas-fcos, efficientdet, spinenet, sp-nas, detnas} focus on searching for a specific module while~\cite{hit-det} directly search for the whole architecture detector.
The automatically designed architectures usually yield higher accuracy or efficiency.

\begin{figure}[t]
    \begin{center}
       \includegraphics[width=0.9\linewidth]{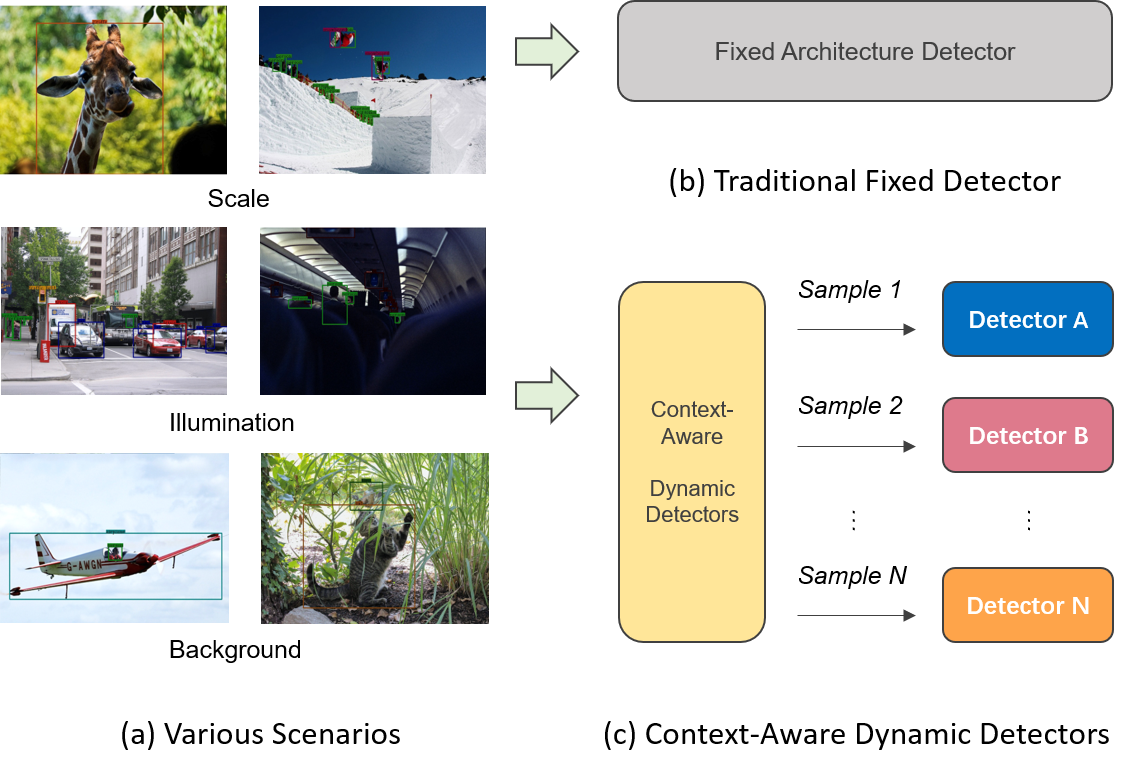}
    \end{center}
       \vspace{-1em}
       \caption{(a). Different input images are with various scenarios.
       (b). Previous methods use a fixed detector for all the inputs.
       (c). CADDet measures the scene complexity and generate a customized architecture for each sample.}
    \label{fig:scene_diversity}
\end{figure}

Previous methods aim to train a generic while fixed detector, \ie, 
they treat all input images equally regardless of the scenario. 
While achieving high accuracy under certain computational costs, 
these models lack the ability of changing capacities according to the content of different inputs.
We claim that it is inefficient to use a fixed model to detect all the samples. 
As shown in Fig.~\ref{fig:scene_diversity}, the scale of objects, illumination and background varies a lot among different samples in the real-world dataset.
Ideally, it is more efficient to use a light-weight model for simple cases while a complex model for hard examples.

Motivated by the above observations, we propose to learn to generate content-aware dynamic detectors~(CADDet).
We aim to train dynamic detectors with the following properties. 
i). The detector is able to automatically generate architectures according to different input images.
ii). For the \emph{esay-to-detect} samples, the generated architectures are light-weight, and vice versa.
iii). For the images with similar visual properties, the corresponding architectures are also similar.

We propose to apply the \emph{dynamic routing} strategy to generate such dynamic detectors.
Dynamic routing is an adaptive inference mechanism for deep neural networks. 
During the inference phase, for each sample, only part of the whole network would be activated, resulting in 
dynamic network architectures.
Recently, dynamic routing have been drawing increasing attention many tasks.
Some works apply 
block dropping~\cite{skipnet, blockdrop, hydranets, msdnet, adainfer, improved, codinet} or 
channel pruning~\cite{dynamic-gc, dynamic-lp, mutualnet} strategies for efficiency in image classification.
Most recently, Li~\etal~\cite{dynamic-seg} proposed a multi-path dynamic network to alleviate scale variance in semantic segmentation.
Different from previous methods, this work focuses on utilizing the \emph{detection-related properties} 
and measuring the \emph{scene complexity} to guide the learning of \emph{dynamic object detectors}. 

We start from designing the routing space. Considering the multi-scale property of detectors, we design the 
multi-scale densely connected supernet with the similar architecture of that in segmentation models 
such as Auto-DeepLab~\cite{auto-deeplab} and Li~\etal~\cite{dynamic-seg}.
With this design, CADDet is able to combine the backbone and neck parts. 
Then, we propose a scale encoding method and introduce a coarse-to-fine strategy to guide the learning of dynamic routing 
tailored for object detection.
On the one hand, 
globally, we assign different samples with various computational budgets~\eg,~FLOPs or MAdds, according to their complexities. 
On the other hand, we introduce a \emph{local} path similarity regularization method. 
We evaluate CADDet on the well-known MS-COCO~\cite{coco} dataset. Results demonstrate CADDet achieves 
1.8 higher mAP with 10\% fewer MAdds compared with the vanilla routing strategy. Compared with the architecture with similar 
building blocks, \ie, MobileNet~\cite{mobilenet}+FPN~\cite{fpn} and MobileNetV2~\cite{mobilenetv2}+FPN, CADDet is able to achieve competitive mAP with 42\% fewer MAdds.

\section{Related Work}
\label{related}

\subsection{Object Detection}
Existing object detection models can be roughly divided into two-stage detectors~\cite{fast-rcnn,faster-rcnn,cascade-rcnn,fpn} and one-stage detectors~\cite{yolo,ssd,retinanet,centernet,fcos}.
The two-stage models tend to be more accurate, 
while the one-stage detectors are usually simpler and thus more efficient. 
In this paper, we focus on model efficiency and thus choose to learn dynamic architectures for one-stage object detectors.
Next, we introduce both the hand-crafted and the NAS-based detection architectures.

\subsubsection{Hand-Crafted Architectures}
Due to the huge scale variance of input samples, 
the representation of multi-scale features is the main problem in object detection. 
Earlier works like SSD~\cite{ssd} and Overfeat~\cite{overfeat} 
directly utilize the feature maps from intermediate layers of the backbone 
to perform the subsequent prediction tasks. 
FPN~\cite{fpn} firstly introduces the lateral connection operation and 
proposes a top-down feature aggregation strategy. After that, there are lots of works~\cite{bfp,pafpn} focusing on designing efficient
\emph{necks}. Besides, backbones with recurrent down-sample and up-sample modules~\cite{hourglass, fishnet} are proposed.
HRNet~\cite{hrnet} tackles this problem by introducing a multi-branch backbone, where the nearby feature maps communicate with 
each other through feature fusion.

\subsubsection{Neural Architecture Search}
Recently, Neural Architecture Search~(NAS) has shown promising results on object detection.
Among them, NAS-FPN~\cite{nas-fpn} and Auto-FPN~\cite{auto-fpn}, as the pioneer works, search for architectures of the neck part. 
EfficientDet~\cite{efficientdet}, DetNAS~\cite{detnas}, and SP-NAS~\cite{sp-nas} learn the backbones and combine them with the fixed necks. 
SpineNet~\cite{spinenet} learns to permute and connect the intermediate layers in the backbone.
Most recently, Hit-Det~\cite{hit-det} proposes to search for the whole detector.
The main different between NAS-based detectors and hand-crafted ones is that NAS implicitly or explicitly adds 
an evolution stage to adjust the architecture of the model. However, the evolution stage is performed to 
achieve higher accuracy on the training or validation set. During the inference phase, the model has no ability to adjust 
itself adaptively.

To summarize, both hand-crafted and automatically designed models aims at finding an optimal while fixed model. Different from these 
models, our CADDet can generate sample-adaptive dynamic architectures. 

\begin{figure*}[t]
    \begin{center}
       \includegraphics[width=1.0\linewidth]{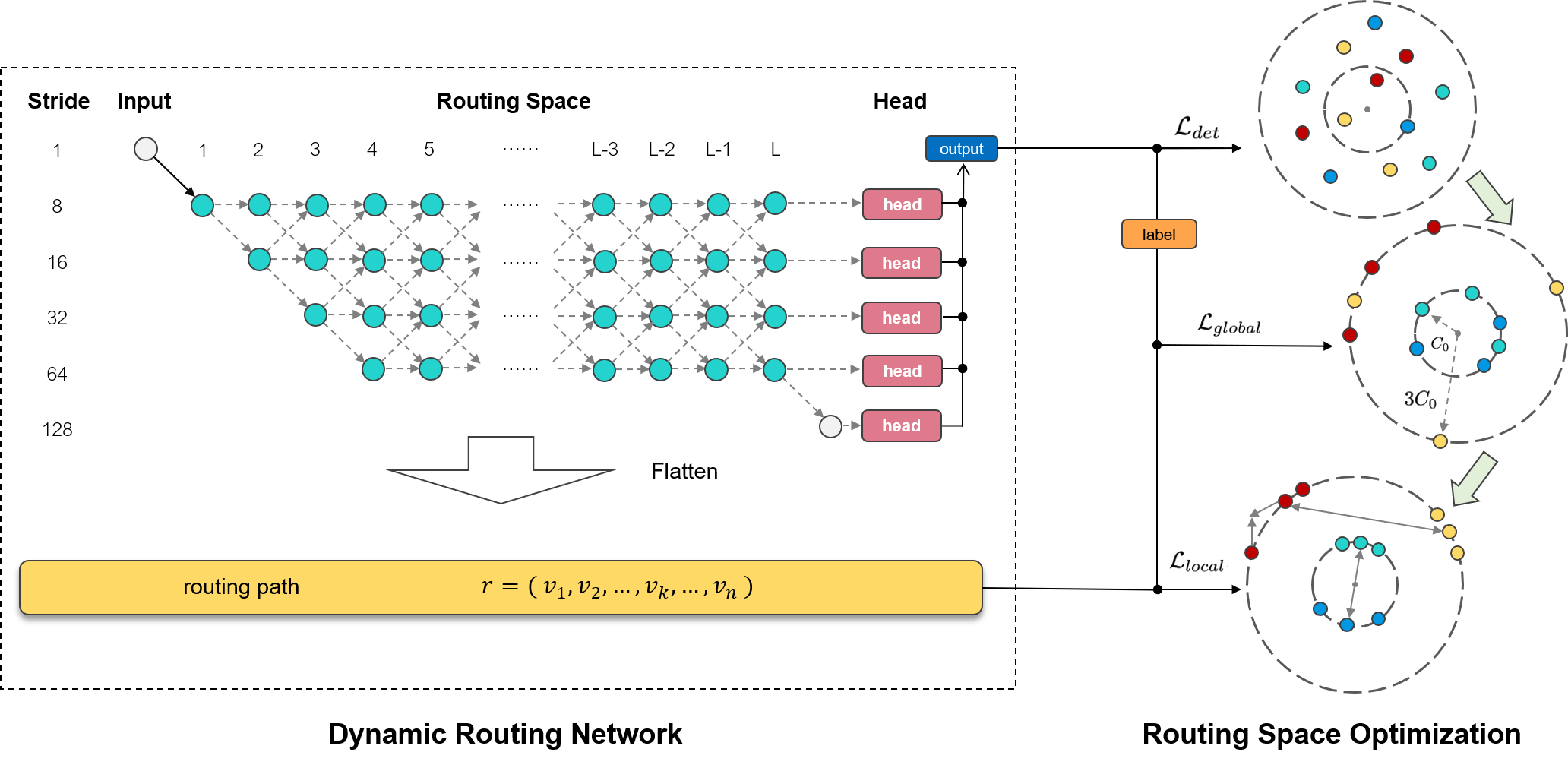}
    \end{center}
        \vspace{-1.5em}
       \caption{ An overview of the proposed CADDet. 
                The left part illustrate the architecture of the multi-scale densely connected network, \ie, supernet.
                The right part shows the optimization in the routing space. 
                We first apply a dynamic global budget constraint~(Sec.~\ref{sec:global}) to regularize the overall budget for each sample.
                Then, we introduce the local path similarity regularization~(Sec.~\ref{sec:local}) to adjust the paths.}
    \label{fig:supernet}
\end{figure*}

\subsection{Dynamic Inference}
In order to achieve higher accuracy or speed up the model, 
various types of dynamic models have been proposed.
During the inference phase, they either generate dynamic model parameters~(dynamic convolution), or 
generate dynamic architectures~(dynamic routing), or perform early prediction depending on the input image.
Among these strategies, our CADDet is most related to dynamic routing.
Dynamic routing models adopt the idea of layer-skipping to generate a suitable sub-network for the current input sample. 
Most of these models are designed for the image classification task.
SkipNet~\cite{skipnet} introduces the recurrent gates to judge whether a certain block would be dropped.
BlockDrop~\cite{blockdrop} utilizes an additional policy network to choose the layers during inference.
HydraNets~\cite{hydranets} increase the width of blocks by adding more specialized components in each layer.
MSDNet~\cite{msdnet} proposes a multi-scale dense network and learns the connection pattern during training.
MutualNet~\cite{mutualnet} trains the supernet with multiple widths using images with multiple input resolutions and introduces 
knowledge distillation during the training process.
CoDiNet~\cite{codinet} proposes to regularize the routes according to sample similarity.
Li~\etal~\cite{dynamic-seg} proposes to learn dynamic routing in semantic segmentation to alleviate the problem of scale variance.
Compared with these methods, our CADDet is the first work to introduce the dynamic routing mechanism in object detection. 
Moreover, we take into consideration the properties of object detection and introduce the global and local regularization 
metrics to guide the learning of dynamic routing.

\section{Content-Aware Dynamic Detectors}
\label{sec:method}

In this section, we introduce the Content-Aware Dynamic Detectors~(CADDet) in detail.
We first outline the overall structure of CADDet as well as the basic dynamic routing mechanism
in Sec.~\ref{sec:dynmaic_detector}.
Next, in Sec.~\ref{sec:global} and Sec.~\ref{sec:local}, 
we introduce a coarse-to-fine regularization strategy tailored for object detection, 
including the dynamic global budget constraint and the local path similarity regularization, 
aiming to 
achieve higher computational efficiency and obtain more diverse architectures.
Finally, we present the architecture details in Sec.~\ref{sec:arc_detail}.

\subsection{Dynamic Routing for Detection}
\label{sec:dynmaic_detector}

In order to increase the capacity of the model and enlarge the routing space, we apply a 
multi-scale densely connected network following MSDNet~\cite{msdnet}, Auto-DeepLab~\cite{auto-deeplab}, and Li~\etal~\cite{dynamic-seg} 
as the supernet. 
As shown in Fig.~\ref{fig:supernet}, the supernet starts with a fixed 3-layer ``stem'' block,
which down-samples the resolution of input images to 1/8.
Then, we add a network which maintains the feature representation at 4 different scales in 
the remaining layers. The 4 scales matches $C3$ to $C6$ respectively as in other detectors.
For the adjacent 2 scales, the feature maps with lower resolutions are with more channels~(2x).
Similar as the design in Li~\etal~\cite{dynamic-seg}, there are 3 candidate paths for each computation node.
The paths include resolution-up, resolution-keep and resolution-down, except for the boundary scales.
Right after the supernet, we concatenate the feature map in each scale with a detection head to generate predictions.
For simplicity, we use the commonly-used one stage dense heads as used in FCOS~\cite{fcos}.
It is worth noting that the supernet itself contains feature maps of multiple scales and there are 
cross-scale feature fusion operations in each node, thus, there is no need to use an additional neck~(\eg, FPN).

\begin{figure*}[t]
    \begin{center}
       \includegraphics[width=0.9\linewidth]{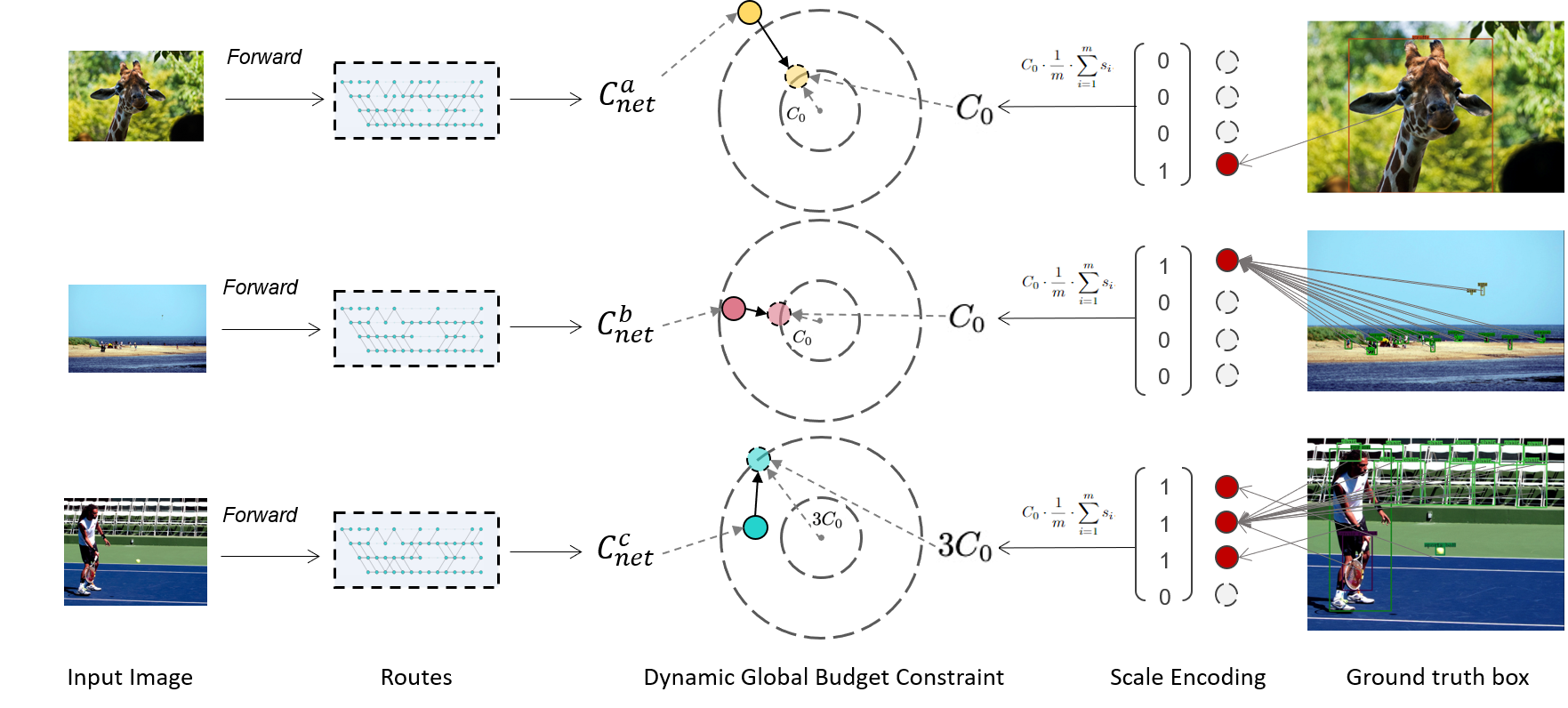}
    \end{center}
       \vspace{-1em}
       \caption{Illustration of scale encoding and global budget constraint}
    \label{fig:global}
\end{figure*}

\begin{figure}[t]
    \begin{center}
       \includegraphics[width=1.0\linewidth]{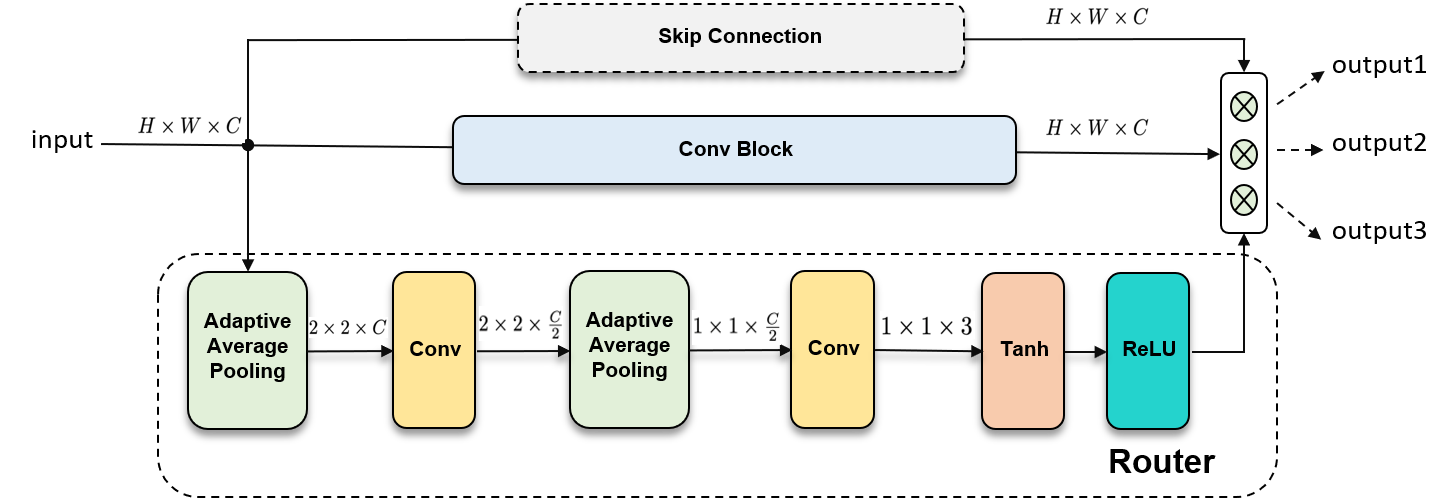}
    \end{center}
       \vspace{-1em}
       \caption{ Illustration of a computation node in our CADDet. }
    \label{fig:router}
\end{figure}

Next, we introduce the router in each computation node. 
As shown in Fig.~\ref{fig:router}, each computation node contains two branches,
\ie, a router and a convolutional block. They take in the same feature map with the shape of
$R^{B\times C\times H\times W}$ as the input, where $B, C, H, W$ represent the batch size, the channel number,
the height and the width.
A router contains a series of Pooling-Conv operations as illustrated in Fig.~\ref{fig:router}, and
outputs a batch of 3-dimentional gate values~$G = {(g^1, g^2, g^3)}^T$, where each element in $G$ represents 
the gate of the corresponding candidate output path~\ie, resolution-up, resolution-keep and resolution-down.
Following Li~\etal~\cite{dynamic-seg}, we allow multi-path propagation, \ie, there would be multiple open gates simultaneously.
During training, the values in $G$ take continuous value for easier back propagation.
In the inference phase, we binarize $G$ by comparing the values with a constant threshold~$\tau$.
If $g^i < \tau$, the corresponding path would be dropped. If all the paths in the current computation node are dropped, the current
convolutional block would also be dropped.
The routers generate different gates for different input samples, which brings about various model architectures.
More details of the network and routers are described in Sec.~\ref{sec:arc_detail}.

\subsection{Dynamic Global Budget Constraint}
\label{sec:global}
To save the computational cost at inference time,
Previous dynamic routing approaches~\cite{dynamic-seg,codinet} add a budget constraint term
in the loss function during the search/routing procedure.
We also introduce the budget constraint in CADDet.
First, we measure the cost~(MAdds) in the $i^{th}$ computation nodes as
\begin{equation}
    \begin{split}
        c_i &=  max(G_i)\cdot c_{conv} + {G_i}^T {[ c_{up}, c_{keep}, c_{down}]}^T,
    \end{split}
    \label{eq:node_cost}			
\end{equation}
where $G^i$ is the gate of the $i^{th}$ node, $c_{conv}$, $c_{up}$, $c_{keep}$, $c_{down}$ represent the 
MAdds (constant values) of the current convolution operation and the resolution change operations respectively.
Let $n$ denotes the total number of the nodes. 
Then, the computational cost of the current architecture can be represented by Eq.~\ref{eq:tot_cost}.
\begin{equation}
    \begin{split}
        C_{net} &= \sum_{i=1}^n c_i.
    \end{split}
    \label{eq:tot_cost}			
\end{equation}
With the total cost $C_{net}$, previous dynamic routing methods either directly minimize C or add an L2 loss
to minimize the gap between $C_{net}$ and an expected MAdds number. 
We claim this is sub-optimal to use a fixed budget constraint for all the inputs since the sample complexity varies a lot and different inputs 
may require different computational resources. 
Therefore, in CADDet, we take a further step by measuring the complexity of difference samples and assign dynamic budget constraints accordingly. 
We introduce a metric to measure the image complexity tailored for object detection 
as described in the following part.


\begin{figure*}[t]
    \begin{center}
       \includegraphics[width=0.85\linewidth]{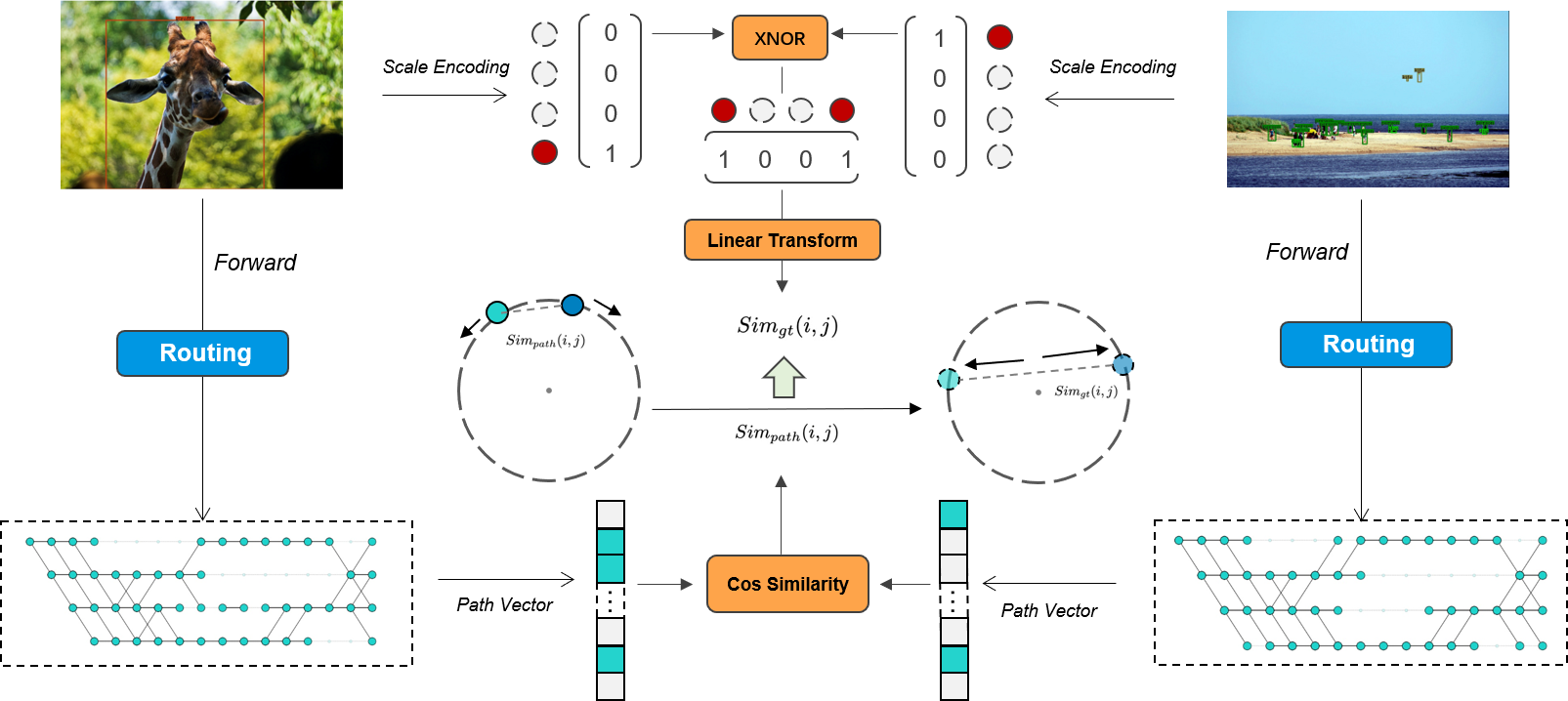}
    \end{center}
        \vspace{-0.8em}
       \caption{Illustration of the local path similarity regularization.}
    \label{fig:local}
\end{figure*}

The scale variance is one of the main challenges in object detection. 
In CADDet, we use the quantized scale distribution to formulate the complexity of each input image.
As shown in Fig.~\ref{fig:global}, we divide the scale space into m intervals. 
Then, we count the shapes of groudtruth bounding boxes in a training image and
generate an m-dimentional scale-encoding vector $S = {(s_1, \cdots, s_m)}^T$.
Each element in $S$ takes binary values, and the $i^{th}$ element in the vector indicates 
whether there exists object(s) with the shape in the $i^{th}$ scale interval.
Finally, we map the scale vector to a dynamic computation budget.
For simplicity, we use the mapping function as Eq.~\ref{eq:map_func},
\begin{equation}
    \begin{split}
        C_{expect} = C_0\cdot\frac{1}{m}\cdot \sum_{i=1}^m s_i,
    \end{split}
    \label{eq:map_func}			
\end{equation}
where $C_0$ is a constant indicating the base MAdds. 
After that, we apply an L2 loss shown in Eq.~\ref{eq:global_constraint} as the dynamic global budget constraint.
\begin{equation}
    \begin{split}
        \mathcal{L}_{global} &= {||C_{net} - C_{expect}||}^2_2.
    \end{split}
    \label{eq:global_constraint}			
\end{equation}
Eq.~\ref{eq:map_func} and Eq.~\ref{eq:global_constraint}
imply that an image with single-scale object(s) is expected to have a light-weight architecture 
while the budget for a complex sample is relaxed. 
Our dynamic global budget fully utilize the scale property as the prior knowledge and experiments
in Sec.~\ref{tab:global_budget} demonstrate the design brings about not only more diverse architectures 
but also lower computational cost and higher accuracy.

\subsection{Local Path Similarity Regularization}
\label{sec:local}
Although the dynamic budget constraint can improve the diversity of routes, it lacks the supervision
of the local structure. For example, the budget constraints for the first two samples in Fig.~\ref{fig:global} are the same since they both contain only one scale of object(s).
However, the scales of these two samples are different. 
Thus, the corresponding paths should differ from each other to some degree.
Motivated by this, we introduce the local path similarity regularization.
For a batch of training samples, we encourage the router 
to generate similar architectures for images with similar scale distributions. 
Conversely, we push away the distance of routes between samples with different scale encoding vectors.

The process of local path similarity regularization is shown in Fig.~\ref{fig:local}. Suppose the batch size is $B$.
For each sample in the batch, we first compute the scale encoding vector $S$ as introduced in Sec.~\ref{sec:global}.
Then, for each pair of samples~$(i, j)$ in the batch~$(1\leq i<j\leq B)$, we use the element-wise
XNOR operation to compute the scale similarity as Eq.~\ref{eq:scale_sim}
\begin{equation}
    \begin{split}
        Sim_{scale}(i,j) = \frac{1}m\cdot \mathrm{sum}(S_i \ \text{XNOR} \  S_j).
    \end{split}
    \label{eq:scale_sim}			
\end{equation}
Next, for each input sample, we flatten the gates of all the computation nodes to form a corresponding
route vector $R_i$. For each pair, we represent the path similarity as Eq.~\ref{eq:route_sim}.
\begin{equation}
    \begin{split}
        Sim_{path}(i,j) &= \cos(R_i, R_j) \\
                        &= \frac{{R_i}^T R_j}{||R_i||\cdot||R_j||}.
    \end{split}
    \label{eq:route_sim}			
\end{equation}
Finally, we compute the path similarity loss by Eq.~\ref{eq:local_constraint}.
\begin{equation}
    \begin{split}
        \mathcal{L}_{local} = \frac1B\cdot \sum_{1\leq i < j \leq B} || Sim_{path}(i,j) - Sim_{gt}(i,j) ||^2_2,
    \end{split}
    \label{eq:local_constraint}			
\end{equation}
where $Sim_{gt}(i,j)$ is linearly transformed from $Sim_{scale}(i,j)$ as Eq.~\ref{eq:linear_transform}, 
which changes the groundtruth similarity from the range of [0,1] to $[Min, Max]$, (0$<$Min$\leq$ Max $\leq$ 1).
\begin{equation}
    \begin{split}
        Sim_{gt}(i, j) = Sim_{scale}(i,j) \cdot (Max-Min) + Min ,
    \end{split}
    \label{eq:linear_transform}			
\end{equation}
Such a transformation allows a part of the architecture to be generic for all the samples.

\begin{table*}[t]
    \centering
    
\begin{tabular}{lcccccc}
    \toprule
    Budget Constraint           & $C_0/C_{tot}$ & mAP   & Avg MAdds~(M) & Max MAdds~(M) & Min MAdds~(M) & MAdds std \\
    \midrule
    \multirow{2}{*}{Fixed}      &     0.1       & 24.7  & 4210      &   4470    & 3969      & 63.7      \\
                                &     0.2       & 25.5  & 4337      &   4572    & 4111      & 76.5      \\
    \midrule
    \multirow{2}{*}{Loss-Aware} &     0.05      & 25.7  & 4174      &   4421    & 3917      & 83.5      \\
                                &     0.1       & 25.9  & 4221      &   4462    & 3997      & 87.2      \\
    \midrule
    \multirow{2}{*}{Proposed}   &     0.05      & 26.2  & \textbf{3912}      &   4492    & \textbf{3458}      & \textbf{147.0}      \\
                                &     0.1       & \textbf{26.4}  & 4007      &   4511    & 3564      & 135.3      \\
    \bottomrule    
\end{tabular}
    \vspace{0.8em}
    \caption{Comparison of different budget constraints. Results are evaluated on COCO validation set.
    MAdds are measured with the input resolution of (640, 800).
    ``MAdds std'' is  the standard deviation term indicating the diversity of the generated routes.
    Note that the ``Fixed'' strategy uses a fixed $C_0$ while the ``Loss-Aware'' and the proposed strategy 
    use a dynamic $C\in [C_0, 4C_0]$. Thus, we take different values of the ratio term for a fair comparison.
    }
    \label{tab:global_budget}
\end{table*}

\subsection{Architecture Details}
\label{sec:arc_detail}
In this section, we introduce the implementation details of CADDet.
The ``stem'' block contains three stride-2 3x3 depth-wise separable convolutions, 
which downsamples the resolution to 1/8~(the scale of C3 in object detection). 
Following Li~\etal~\cite{dynamic-seg}, we set the layer number of the supernet to 16 by default.
The four scales of feature maps are with the channel numbers of 64, 128, 256, 512 respectively.

In each computation node of scale $s$, 
the input feature map is the summation of the outputs from the nodes of the previous layer 
with scales of $s/2, s, 2s$. As shown in Fig.~\ref{fig:router}, the router first
utilizes an average pooling operation, which downsamples the feature map to 2x2.
Then, a stride-1 1x1 convolution, a global average pooling~(GAP) layer and a fully-convolutional layer 
are cascaded. The router finally outputs a tensor with the shape of $(B,3)$, representing 
the gates for the current $B$ samples. The hyper-parameter for gate discretization $\tau$ is set to $1e-4$
As for the convolutional block, we applied the commonly used 3x3 SepConv for higher efficiency.
The resolution-up operation is a stride-1 1x1 convolution with bilinear interpolation.
The resolution-down operation is a stride-2 1x1 convolution. The 1x1 convolutions are applied 
to align the channel numbers between adjacent scales.

The supernet is connected with the detection head~\ie, FCOS head~\cite{fcos} 
without an additional neck. Specifically, the feature maps from C3 to C6 are directly obtained 
by projecting the 4 output feature maps to 256-d. C7 is obtained by adding a convolution layer on C6.

Scale encoding is the key procedure for both the global budget constraint and the local path similarity regularization.
We analyze the scale distributions of samples on MS-COCO 
quantize the scales by the 4 intervals, [0, 64], (64, 150], (150, 360] and (360, $\infty$).
During the scale encoding process, the scale of each object is determined by its longer side, \ie, $\max(h,w)$.
The base MAdds $C_0$ is set to $5\%$ of the total MAdds of the supernet.
For the local path similarity regularization, we set $Min$ and $Max$ to 0.6 and 0.95 respectively.

Putting all the loss terms together, we can obtain the overall objective function as Eq.~\ref{eq:tot_loss},
\begin{equation}
    \begin{split}
        \mathcal{L}_{tot} = \mathcal{L}_{det} + \lambda_1 \mathcal{L}_{global} + \lambda_2 \mathcal{L}_{local},
    \end{split}
    \label{eq:tot_loss}			
\end{equation}
where $\mathcal{L}_{det}$ is the detection loss, including the losses for classification and localization.
$\lambda_1$ and $\lambda_2$ are the hyper-parameters to balance the respective losses, 
which take $1.0$ as the default value.

\section{Experiments}
\label{sec:experiment}

\subsection{Experimental Setup}
\label{sec:set_up}
In this section, we introduce the implementation details of the training process.
We first pre-train our supernet on the ImageNet dataset~\cite{imagenet}~(ILSVRC2012). We use the SGD optimizer 
with 0.2 as the initial learning rate. The batch size is set to 256 and the model is trained for 120 epochs.
During pre-training, we remove the routers and all the gate values are set to 1.

Then, we conduct the main experiments on the MS-COCO~\cite{coco} dataset. Following the common practice, 
we use the COCO \textit{trainval35k} split~(115K images) for training and \textit{minival} 
split~(5K images) for validation. We also report the main results on the \textit{test\_dev} split~(20K images).
We train the CADDet with the initial learning rate being $0.01$. Batch size $B$ is set to 16.
Unless specified, we apply the $1x$ schedule. The model is trained for 12 epochs and the 
learning rate is reduced by 10 after the $8^{th}$ and $11^{th}$ epochs.
Besides, 
neither $\mathcal{L}_{local}$ nor $\mathcal{L}_{global}$ is applied during the first epoch
to avoid model degeneration.
In our ablation study, we resize the images such that the shorter side is 600 and the longer side
is less than or equal to 1000. In the main results, the images are resized to have their shorter side 
being 800 and the longer side is less than or equal to 1333. By default, we use the FCOS head as the 
detection head. All of our experiments are conducted using mmdetection~\cite{mmdetection} 2.1.

\subsection{Ablation Study}
\label{sec:ablation}

\subsubsection{Different Global Budget Constraints}
\label{sec:ablation_global}
In this section, we study the effectiveness of different budget constraints.
We apply the following 3 different kinds of budget constraints.

i). Fixed budget constraint. All the samples are assigned with the same budget constraint, which
is commonly used in other NAS and other dynamic routing methods. This method serves as 
the baseline global budget constraint.

ii). Loss-aware budget constraint. During training, we create a buffer to hold the detection losses of the least recently used 
100 input samples. Next, for the current training sample, we first compute the detection loss and find the rank of 
it in the buffer. Then, we linearly transform the rank to an expected budget in the range of $[C_0, 4C_0]$.

iii). The proposed dynamic global budget constraint, which utilizes the scale property as the 
prior information to measure the sample complexity.

The quantitative results are shown in Table~\ref{tab:global_budget}. According to the results, 
all these budget constraints help to generate different models for different inputs.
Compared with the fixed budget constraint, the loss-aware and the proposed constraints can generate more efficient architectures~\ie, 
these approaches can achieve similar accuracy with relatively lower cost. 
Compared with the loss-aware approach, the proposed scale-aware strategy can generate more diverse routes. The high diversity also 
brings about 1.5 to 2.0 p.p. higher mAP with lower average MAdds. These demonstrate scale encoding is 
a proper measure of sample complexity for object detection.

\begin{table}[t]
    \centering
    
\begin{tabular}{lccccc}
    \toprule
    Method                      & Min & Max &  mAP   & MAdds~(M)   &  std \\
    \midrule
    Global                      & -   & -   &  26.2     & 3912     & 147.0  \\
    \midrule
    \multirow{5}{*}{+Local}      & 0.4 & 0.95&  24.3  & 3753     & \textbf{211.3}  \\
                                & 0.6 & 0.95 &  \textbf{26.5}  & \textbf{3725}     & 195.3  \\
                                & 0.8 & 0.95&  26.1  & 3901     & 107.7  \\ 
                                & 0.6 & 0.9 &  26.3  & 4011     & 170.0  \\
                                & 0.6 & 1.0 &  26.2  & 3951     & 95.8  \\
    \bottomrule    
\end{tabular}
    \vspace{0.8em}
    \caption{Ablation for the local constraint. Results are evaluated on the COCO validation set.
    MAdds are measured with the input resolution of (640, 800).
    The $Min$ and $Max$ are hyper-parameters as described in Sec.~\ref{sec:local}.
    }
    \label{tab:local_constraint}
\end{table}

\subsubsection{Using the Local Path Similarity Regularization}
\label{sec:ablation_local}

\begin{figure*}[t]
    \begin{center}
       \includegraphics[width=0.9\linewidth]{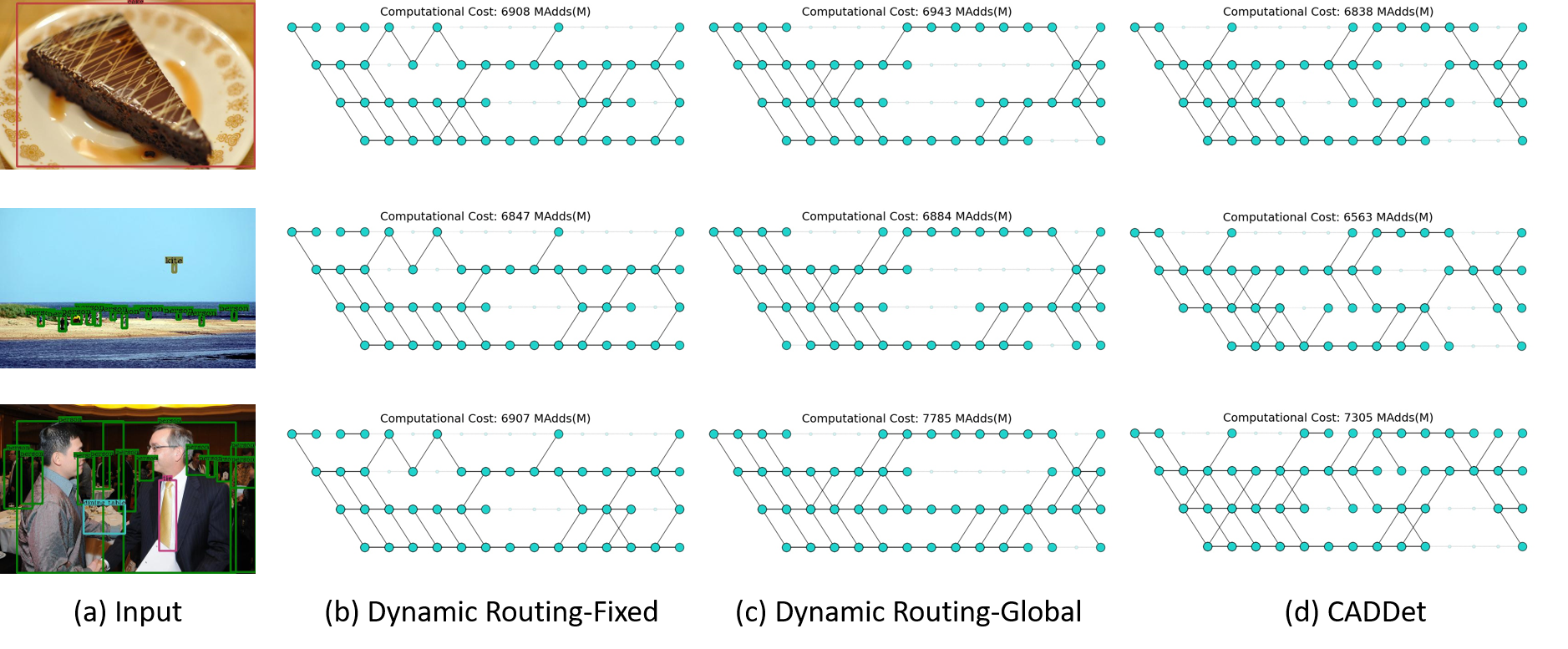}
    \end{center}
       \vspace{-1.5em}
       \caption{Qualitative results on the validation set of COCO. MAdds are measured with the input resolution of (800, 1200).}
    \label{fig:qualitative}
\end{figure*}

In this section, we study the effectiveness of the proposed local path similarity regularization.
Quantitative results are listed in Table~\ref{tab:local_constraint}. 
We can find adding the local constraint can further improve the diversity of routes.
We also study the effectiveness of hyper-parameters $Min$ and $Max$. As introduced in Sec.~\ref{sec:local},
these two parameters are designed to control the lower and upper bounds of groudtruth for the path similarity.
According to the results in Table~\ref{tab:local_constraint}, a small lower bound results in worse 
accuracy, while a larger value leads to higher computational cost. This implies it is better to 
allow the routers to generate a generic architecture while let them to have a certain degree of freedom.
The comparison of routes generated by different strategies can be found in Fig.~\ref{fig:qualitative}.
The analysis could be found in Sec.~\ref{sec:qualitative}.

\begin{table}[t]
    \centering
    
\begin{tabular}{lcccc}
    \toprule
    Model structure             & $C_0/C_{tot}$ &  mAP   & MAdds~(M)\\
    \midrule
    \multirow{2}{*}{CADDet}     & 0.05&  26.5  & 3725     \\
                                & 0.1 &  26.4  & 3825     \\
    \midrule
    \multirow{2}{*}{CADDet+FPNLite} & 0.05&  26.3  & 3952 \\
                                & 0.1 &  26.8  & 4119     \\
    \bottomrule    
\end{tabular}
    \vspace{0.8em}
    \caption{
        Comparison of different detection structures. According to the results, the absence of FPNLite
        would not affect the accuracy and cost. For clearer a comparison, we count the MAdds 
        of the backbone and neck.
    }
    \label{tab:ablation_fpn}
\end{table}

\begin{figure}[t]
    \begin{center}
       \includegraphics[width=0.9\linewidth]{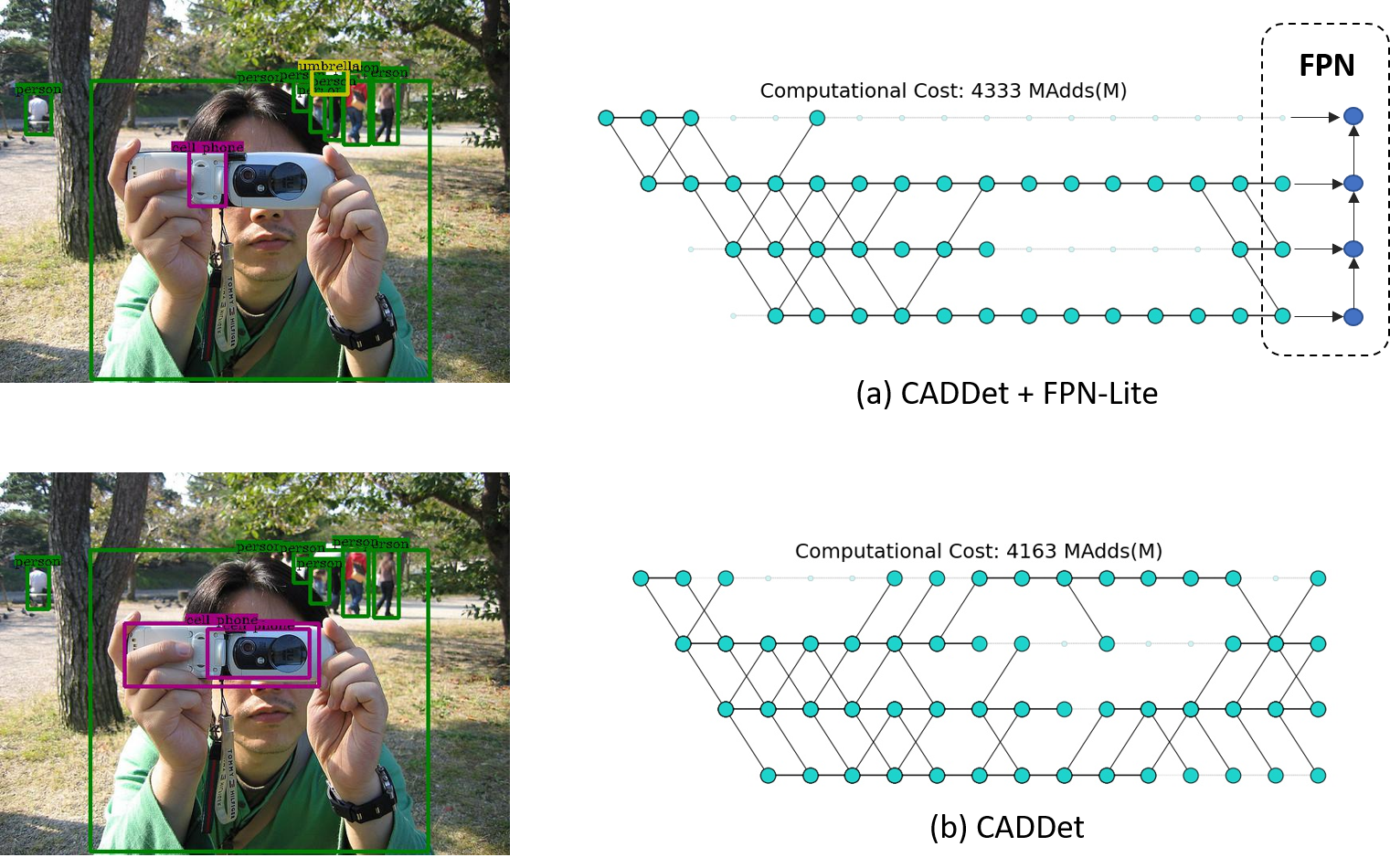}
    \end{center}
       \vspace{-1em}
       \caption{Illustration of routes for (a). \textit{CADDet+FPNLite} and (b). \textit{CADDet}. 
       We compute MAdds of both the backbone and the neck parts. Inputs are with the resolution of (640, 800)}
    \label{fig:ablation_fpn}
\end{figure}

\subsubsection{Using FPNLite}
\label{sec:ablation_fpn}
In Sec.~\ref{sec:dynmaic_detector}, we claim that the supernet itself can extract multi-scale features, thus,
CADDet no longer requires an additional neck to generate multi-scale feature maps. 
In this section, we study the ability of multi-scale feature extraction of the supernet.
We train the following two types of detectors. i) \textit{supernet+FPNLite+head}, 2) \textit{supernet+head}.
From Table~\ref{tab:ablation_fpn}, we find the capacity of our supernet is high enough to represent 
multi-scale features. With lower computational cost, our supernet alone can achieve comparable accuracy.
The routes are illustrated in Fig.~\ref{fig:ablation_fpn}.
It is interesting that without the explicit neck, the routers are able to learn the ``U-shape''
cross-scale feature aggregation operations.

\begin{table*}[t]
    \centering
    
\begin{tabular}{lccccccc}
    \toprule
    Backbone and Neck              & Head                           & Eval Set     & mAP   & Avg MAdds~(G) & Max MAdds~(G) & Min MAdds~(G) \\
    \midrule
    \multirow{4}{*}{MobileNet+FPNLite} & \multirow{2}{*}{FCOS}      &     val      & 25.2  &  11.8 &   \multirow{4}{*}{-}    & \multirow{4}{*}{-}      \\
                                   &                                &     test     & 25.5  &  11.8 &       &       \\
                                   & \multirow{2}{*}{RetinaNet}     &     val      & 25.1  &  11.8 &       &       \\
                                   &                                &     test     & 25.7  &  11.8 &       &       \\
    \midrule
    \multirow{4}{*}{MobileNetV2+FPNLite} & \multirow{2}{*}{FCOS}    &     val      & 28.5  &  7.8  &   \multirow{4}{*}{-}    & \multirow{4}{*}{-}      \\
                                   &                                &     test     & 28.2  &  7.8  &       &       \\
                                   & \multirow{2}{*}{RetinaNet}     &     val      & 24.7  &  7.8  &       &       \\
                                   &                                &     test     & 24.9  &  7.8  &       &       \\
    \midrule
    \multirow{4}{*}{CADDet} & \multirow{2}{*}{FCOS}                 &     val      & 28.5  &  6.5  &   7.6    & 6.0      \\
                                   &                                &     test     & 28.9  &  6.4  &   7.7    & 5.9      \\
                                   & \multirow{2}{*}{RetinaNet}     &     val      & 27.1  &  6.7  &   8.0    & 6.3      \\
                                   &                                &     test     & 27.5  &  6.6  &   8.0    & 6.2      \\
    \bottomrule    
\end{tabular}
    \vspace{0.5em}

    \caption{Main results on COCO \textit{test-dev}.
    The accuracy is measured with the resolution of (800, 1333) and MAdds are measured with the input resolution of (800, 1200).
    Note that we only count  \textbf{MAdds for the backbones and necks}.
    We compare difference kinds of efficient backbones, cascaded with the commonly used FCOS head and 
    Retina head. 
    }
    \label{tab:global_budget}
\end{table*}

\subsection{Qualitative Results}
\label{sec:qualitative}

In this section, we visualize the model architectures generated by different routing strategies.
Results are illustrated in Fig.~\ref{fig:qualitative}. In this part, we compute MAdds under the resolution of (800, 1200).
We have the following observations.

i). Both the global budget constraint and the local path similarity regularization can improve the diversity of 
routes. The main difference is that the local path similarity regularization focuses more on the detailed parts 
rather than the body architecture.

ii). Universal body part.
Our CADDet tends to generate a generic architecture as the body part and adjusts some local 
parts according to different inputs. The body part of most routes is a ``U-shape'' structure that 
first down-samples the feature maps in the \text{head} part and then up-samples in the ``tail'' part.
This pattern is similar with the hand-crafted encoder-decoder structure.

iii). Diversity of CADDet.
The computation nodes in the mid-layers are more diverse compared with the head-layers and tail-layers.
For inputs with small objects, the number of valid computation nodes in the high-resolution feature maps is larger.

\subsection{Main Results}
\label{sec:qualitative}
Finally, we report the results on COCO \textit{test-dev}. For a fair comparison, we compare 
CADDet with other light-weight backbones~\ie, MobileNet and MobileNetV2 which utilize SepConv as the building block.
Results demonstrate that our CADDet can achieve competitive accuracy with much lower computational cost.
\section{Conclusion}
\label{sec:conclusion}

In this paper, we propose to improve the efficiency of detectors by learning
to generate content-aware dynamic detectors~(CADDet).
Specifically, we propose a scale encoding method to i) measure the global sample complexity and 
assign the corresponding budget constraint, and ii) regularize the local similarity of the paths between 
different samples. Experimental results demonstrate CADDet can significantly improve the model diversity 
and save the computational cost. In the future, we aim to learn high-performance dynamic detectors
by utilizing other building blocks.
It is also interesting to generate dynamic heads, where the detection head is also involved in 
the dynamic framework.

{\small
\bibliographystyle{ieee_fullname}
\bibliography{egbib}
}

\end{document}